%% file: IJCNN_main.tex
\documentclass[conference]{IEEEtran}
\IEEEoverridecommandlockouts
\usepackage{cite}
\usepackage{amsmath,amssymb,amsfonts}
\usepackage{algorithmic}
\usepackage{graphicx}
\usepackage{textcomp}
\usepackage{xcolor}

\usepackage{bm,algorithmic,textcomp,booktabs,xcolor}
\usepackage{pgfplots,tikzscale}
\usepackage{amsmath,amsthm}
\usepackage{amsfonts}
\usepackage{pgfgantt}
\usepackage{soul}
\usepackage{makecell}
\usepackage{amsmath, amsthm, amsfonts, amssymb}
\graphicspath{{./fig/}}
\DeclareMathOperator*{\argmin}{argmin}

\def\BibTeX{{\rm B\kern-.05em{\sc i\kern-.025em b}\kern-.08em
    T\kern-.1667em\lower.7ex\hbox{E}\kern-.125emX}}

\makeatletter
\newcommand{\linebreakand}{%
\end{@IEEEauthorhalign}
\hfill\mbox{}\par
\mbox{}\hfill\begin{@IEEEauthorhalign}
}
\makeatother

\begin{document}

\title{Feature Selection Using Batch-Wise Attenuation and Feature Mask Normalization
\thanks{This research was supported by Advantest as part of the Graduate School ``Intelligent Methods for Test and Reliability'' (GS-IMTR) at the University of Stuttgart.}}

\author{\IEEEauthorblockN{Yiwen Liao}
\IEEEauthorblockA{\textit{Institute of Signal Processing and} \\
\textit{System Theory, University of Stuttgart}\\
Stuttgart, Germany\\
yiwen.liao@iss.uni-stuttgart.de}
\and
\IEEEauthorblockN{Rapha\"el Latty}
\IEEEauthorblockA{\textit{Applied Research and Venture Team} \\
\textit{Advantest Europe GmbH}\\
B\"oblingen, Germany \\
raphael.latty@advantest.com}
\and
\IEEEauthorblockN{Bin Yang}
\IEEEauthorblockA{\textit{Institute of Signal Processing and} \\
\textit{System Theory, University of Stuttgart}\\
Stuttgart, Germany \\
bin.yang@iss.uni-stuttgart.de}
}

\maketitle

\begin{abstract}
	Feature selection is generally used as one of the most important preprocessing techniques in machine learning, as it helps to reduce the dimensionality of data and assists researchers and practitioners in understanding data. Thereby, by utilizing feature selection, better performance and reduced computational consumption, memory complexity and even data amount can be expected. Although there exist approaches leveraging the power of deep neural networks to carry out feature selection, many of them often suffer from sensitive hyperparameters. This paper proposes a feature mask module (FM-module) for feature selection based on a novel batch-wise attenuation and feature mask normalization. The proposed method is almost free from hyperparameters and can be easily integrated into common neural networks as an embedded feature selection method. Experiments on popular image, text and speech datasets have shown that our approach is easy to use and has superior performance in comparison with other state-of-the-art deep-learning-based feature selection methods.
\end{abstract}


\input{intro}
\input{formulation}
\input{method}
\input{exp}

\input{conclusion}

\bibliographystyle{unsrt}
\bibliography{./Refs.bib}

\end{document}

%% file: intro.tex
\section{Introduction}
\label{sec:intro}
Feature selection is an important research field in the machine learning community. Typically, given a dataset consisting of samples with $D$ features, feature selection aims to select $K$ from the $D$ features with $K<D$. Subsequently, samples consisting of the selected features are used as a new dataset to solve some given learning tasks, such as classification or regression~\cite{li2017feature}. In this paper, we refer to the learning tasks performed with the selected $K$ features as downstream tasks. Thereby, feature selection helps to reduce data dimensions for a more robust performance in downstream learning tasks. Furthermore, feature selection can be used by researchers and practitioners to find out which features are critical for a given task. Specifically, in some fields such as gene analysis~\cite{li2016deep}, feature selection is used to analyze measurements and find out the salient quantities. 

Due to its importance, feature selection is well established in the literature, especially in the field of conventional machine learning~\cite{li2017feature,guyon2003introduction}. Early feature selection methods (e.g. correlation analysis~\cite{guyon2003introduction} and information-theory-based approaches~\cite{brown2012conditional,peng2005feature}) typically rely on statistical properties of data. However, they usually focus on the relation between single features and the target label, or focus on the features themselves, failing to leverage the possibly available label information. Moreover, complex relations might be neglected in conventional feature selection methods, because they typically presume simple linear relations among different features and labels.
\begin{figure*}[!ht]
	\centering
	\includegraphics[width=0.9\linewidth]{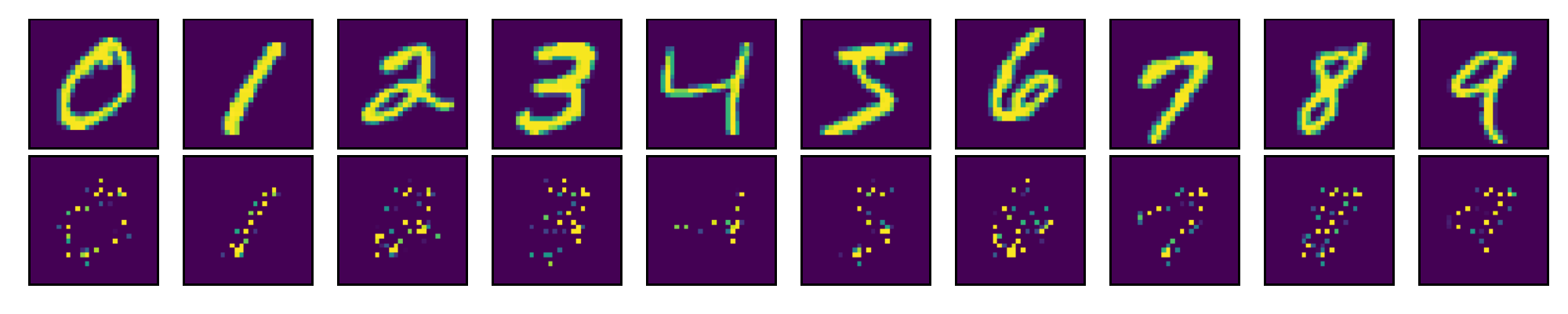}
	\caption{An exemplary feature selection result of the FM-module on MNIST. (\emph{top}) The original features (pixels) of the image data. (\emph{bottom}) The top-50 features selected by the FM-module, corresponding to only about 3.2\% of all pixels, enable more than 95\% classification accuracy with a random forest classifier.}
	\label{fig:mnist_fm_example}
\end{figure*}

Recently, deep learning has achieved dramatic success in many machine learning fields~\cite{goodfellow2016deep}. Naturally, designing feature selection methods using deep learning concepts or in combination with neural networks has become a new research hotspot. In that context, many deep-learning-related techniques are leveraged to provide new ideas for feature selection, such as the attention mechanism~\cite{gui2019afs,wang2014attentional} and autoencoder structures~\cite{abid2019concrete,han2018autoencoder}. Moreover, due to the nature of the non-linearity of deep neural networks, more efficient and robust embedded feature selection methods can be expected if neural networks are used as backbone algorithms. For example, several popular deep-learning-based embedded feature selection approaches such as~\cite{borisov2019cancelout,trelin2020binary,li2016deep} integrate the sparsity penalty from conventional feature selection approaches into neural networks. Nonetheless, existing deep-learning-based feature selection methods still face multiple challenges. First, many of them often necessitate special loss terms in addition to the original learning objective. Thereby, the design of loss terms and the search for a good combination of hyperparameters become new bottlenecks for these methods. Second, some methods utilize redundantly large network architectures by assigning a subnetwork (e.g., an attention network) to each individual input feature. This is usually infeasible for high-dimensional data, which are, however, the most common data in feature selection tasks. Last but not least, many existing deep-learning-based feature selection methods often neglect the relations between features; i.e., features are considered more or less independently during training. This might lead to suboptimal feature selection performance.

In order to address the issues mentioned above and provide new insights of deep-learning-based feature selection, we propose a feature mask module (FM-module) for feature selection based on a novel batch-wise attenuation and feature mask normalization. The FM-module can be jointly trained with common neural networks, such as multiple layer perceptrons (MLP) or convolutional neural networks (CNN)~\cite{goodfellow2016deep}. Once the entire model is trained, the FM-module can directly output a feature mask (FM), of which each element represents the importance score of the corresponding feature of the input data. As a result, the importance scores can be used to select the most representative and informative feature subset. Randomly selected exemplary results on the MNIST dataset~\cite{lecun1998gradient} are shown in Fig.~\ref{fig:mnist_fm_example} to demonstrate the superior performance of the proposed method. In a nutshell, the major contributions of this work can be summarized as follows:

\begin{itemize}
	\item A novel feature mask module is proposed, which can be jointly trained with an arbitrary neural network. The trained FM-module can directly generate feature mask (importance scores) to perform feature selection.
	\item The relation between the original features is taken into account by applying a novel feature mask normalization. Moreover, the batch-wise attenuation forces the same feature mask for all samples within a training batch during each iteration. This targets better to feature selection than the conventional sample-wise attention mechanism.
	\item The proposed method does not introduce additional loss terms to the original learning objectives in comparison with other existing deep learning based approaches, and it is thus easy to use.
	\item The features selected by the FM-module are reliable and have shown similar performance in the downstream tasks with different learning algorithms.
\end{itemize}

%% file: formulation.tex
\section{Related Work}
\label{sec:related-work}
According to~\cite{li2017feature,guyon2003introduction}, feature selection methods can be roughly categorized into three groups: wrapper methods, filter methods and embedded methods. In this paper, we only focus on embedded methods, since most deep-learning-based feature selection methods are generally considered as embedded methods. It should be noted that, in our context, ``deep-learning-based'' means that a feature selection method either uses deep neural networks as a part of it, or takes advantage of the techniques developed within the deep learning community. Furthermore, according to~\cite{guyon2003introduction}, feature selection is sometimes called variable selection. In order not to confuse with the random variable in probability theory, we only use the term ``feature selection'' in this paper.

Early deep-learning-based feature selection methods were frequently inspired by conventional embedded methods, i.e., based on additional sparsity penalty terms in the learning objectives, or based on the analysis of the weights in the hidden layers. Representative studies include~\cite{li2016deep} and \cite{roy2015feature}. In~\cite{li2016deep}, the authors proposed Deep Feature Selection (DFS) by utilizing $\ell_1$- and $\ell_2$-regularization simultaneously. Unlike previous methods such as~\cite{zou2005regularization}, the first layer of DFS was innovatively designed as a sparse one-to-one linear layer; i.e., each input feature only connected to one specific neuron in the first hidden layer. This novel design has become a paradigm for many successors such as~\cite{borisov2019cancelout,trelin2020binary}. In~\cite{roy2015feature}, the importance scores of each input feature were obtained by calculating individual contributions of each input feature using the activation potentials of the first hidden layer. This method does not introduce new architectures or loss terms to the network for the original learning task, but its performance significantly depends on the architecture of the neural network and consequently it does not generalize well to different tasks and neural networks.

Another modern research direction is to use attention mechanism to automatically select features. In~\cite{gui2019afs}, the so called attention based feature selection (AFS) was proposed by constructing attention subnetworks to generate a selection probability matrix. Each attention subnetwork outputs one probability whether one input feature should be selected. In addition to the attention mechanism, $\ell_2$-regularization terms were used in AFS as well.

Moreover, to design special layers or modules by considering stochastic properties is a new trend in the feature selection research, such as ~\cite{trelin2020binary,borisov2019cancelout,abid2019concrete}. For example, in concrete autoencoder~\cite{abid2019concrete}, the authors proposed a concrete selection layer by sampling random variables from concrete distributions. By using a temperature annealing schedule, the concrete random variables can smoothly approximate one-hot coding vectors and can thus be used to select features. Analogously, the binary stochastic filtering (BSF)~\cite{trelin2020binary} samples random variables from a Bernoulli distribution to stochastically select features.

\section{Problem Formulation}
\label{sec:formulation}
This section formulates the feature selection problem in the context of embedded selection methods. Given a dataset of $N$ samples $X_\text{all} = [\bm{x}_1, \bm{x}_2, \cdots, \bm{x}_N]^\top\in \mathbb{R}^{N\times D}$, each sample is denoted as $\bm{x}_i = [x_{i, 1}, x_{i, 2}, \dots, x_{i, D}]^\top\in\mathbb{R}^D$. The corresponding ground truth set for $X_\text{all}$ is denoted as $\bm{y} = [y_1, y_2, \dots, y_N]^\top$ if the downstream problem is a supervised learning task. Feature selection aims to select a subset $\mathcal{S}\subsetneqq\{1, 2, \cdots, D\}$ of the original $D$ features with $|\mathcal{S}| = K$ and $K<D$, i.e. to select $K$ from the $D$ columns of the dataset matrix $X_\text{all}$. In addition, $f(\cdot)$ is used to denote the mapping from the original feature space to the selected feature space: $f(\cdot):\mathbb{R}^D\mapsto\mathbb{R}^K$. Consequently, under the embedded feature selection setup, the subset $\mathcal{S}$ is obtained in such a way that the loss function of the downstream learning task is minimized over the data distribution $p_\text{data}$. Formally, the objective is defined as
\begin{equation}
\label{eq:fs}
\argmin_{f,\ g}\mathbb{E}_{(\bm{x}, y)\sim p_\text{data}}\Big[\mathcal{L}\big(g(f(\bm{x})), y)\big)\Big],
\end{equation}
where $\mathcal{L}(\cdot, \cdot)$ is the loss function targeting a given learning task, and $g(\cdot)$ is the corresponding model, e.g., an arbitrary neural network. However, in most cases, the real data distribution $p_\text{data}$ is unknown and we instead minimize the loss defined in~\eqref{eq:fs} over the empirical distribution defined on the available dataset under the assumption that all samples are independent and identically distributed (\emph{i.i.d.}).

Frequently, we do not directly search for the optimal $f(\cdot)$ because the selection procedure itself is typically not differentiable and consequently difficult to optimize. Instead we search for $f(\bm{x};\bm{m}) = \bm{m}\odot\bm{x}$ such as in~\cite{trelin2020binary,borisov2019cancelout,gui2019afs} with  $f(\cdot;\bm{m}):\mathbb{R}^D\mapsto\mathbb{R}^D$. Correspondingly, the elements of the resulting feature mask vector $\bm{m}\in\mathbb{R}_{\ge 0}^D$ can be interpreted as importance scores for each individual feature with respect to a given downstream learning task. By sorting the importance scores in a descending order, the top-$K$ features can then be selected. Note that the mask $\bm{m}$ can be a function of the input $\bm{x}$ as in~\cite{gui2019afs}, while it can also be independent as in~\cite{borisov2019cancelout}. 

%% file: method.tex
\section{Method}
\label{sec:method}

\begin{figure}[!ht]
	\centering
	\includegraphics[height=0.08\textheight]{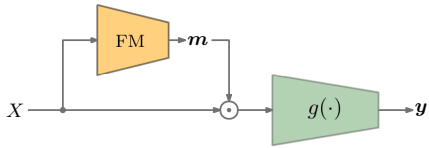}
	\caption{The structure of the proposed feature selection method. The FM-module (\emph{orange}) can generate a fixed feature mask $\bm{m}$ from the data $X$. The learning network $g(\cdot)$ (\emph{green}) can be an arbitrary network parameterized by $\bm{\Theta}$ with respect to a given learning task.}
	\label{fig:network}
\end{figure}

The core idea of the proposed method is the novel batch-wise attenuation within a minibatch during each training iteration, and the feature mask normalization. Thereby, based on these two concepts, we propose a novel \emph{Feature Mask Module} (FM-module) to enable feature selection, which is jointly trained with an arbitrary deep neural network as shown in Fig.~\ref{fig:network}. Once the FM-module is trained on a given dataset, the top-$K$ features can easily be obtained by comparing the importance scores of each individual input feature in the learned feature mask.

\subsection{Network and Objective Function}
\label{subsec:network}

In this section, we present how to use the proposed FM-module in common neural networks. Specifically, let $g(\cdot; \bm{\Theta})$ be an arbitrary neural network for a given learning task parameterized by $\bm{\Theta}$, such as classification or regression, and $\mathcal{L}(\cdot, \cdot)$ denote the corresponding loss function. The given dataset is denoted as $\{(\bm{x}_1, y_1), (\bm{x}_2, y_2), \dots, (\bm{x}_N, y_N)\}$. A common learning task \emph{without} feature selection can be defined as
\begin{equation}
	\label{eq:loss-wo-fs}
	\argmin_{\bm{\Theta}}\mathbb{E}_{(\bm{x}, y)\sim p_\text{data}}\Big[\mathcal{L}\big(g(\bm{x}; \bm{\Theta}), y\big)\Big]
\end{equation}
In order to learn the feature mask vector $\bm{m}$, the FM-module is added before the downstream network $g(\cdot; \bm{\Theta})$ as shown in Fig.~\ref{fig:network}. Thereby, each input for $g(\cdot; \bm{\Theta})$ becomes $f(\bm{x})=\bm{m}\odot\bm{x}_i$, where $\bm{m}$ is dependent on the training batches. Consequently, as an embedded feature selection method, the overall learning objective becomes
\begin{equation}
	\label{eq:loss-w-fs}
	\argmin_{\bm{\Theta}, \bm{\Theta}_\text{FM}}\mathbb{E}_{(\bm{x}, y)\sim p_\text{data}}\Big[\mathcal{L}\big(g(f(\bm{x}; \bm{\Theta}_\text{FM}); \bm{\Theta}), y)\big)\Big]
\end{equation}
where $f(\cdot; \bm{\Theta}_\text{FM})$ denotes the mapping defined by FM-module and the element-wise multiplication, parameterized by $\bm{\Theta}_\text{FM}$.\footnote{In our running example, $\bm{\Theta}_\text{FM} = \{W_1, W_2, \bm{b}_1, \bm{b_2}\}$ defined in Subsection~\ref{subsec:fm-module}.}

What makes the proposed FM-Module appealing is that no additional regularization terms or carefully designed loss terms need to be added to the original downstream learning objective. This means that the entire network shown in Fig.~\ref{fig:network} can be directly trained to minimize exactly the same loss function $\mathcal{L}(\cdot, \cdot)$ for $g(\cdot; \bm{\Theta})$. For example, if the downstream task is classification, a canonical objective function could be the cross entropy loss, while the mean squared error loss can be the objective function for a regression task.

\subsection{Feature Mask Module}
\label{subsec:fm-module}
\begin{figure}
	\centering
	\includegraphics[width=0.75\linewidth]{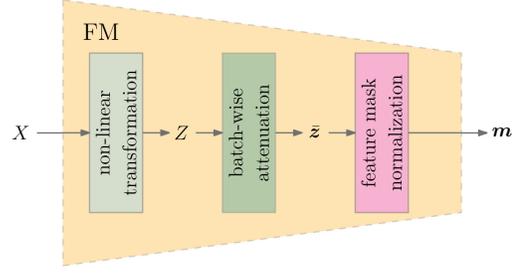}
	\caption{The structure of the FM-module. In each iteration, a minibatch $X$ is mapped to $Z$ under a non-linear transformation. Subsequently, a primitive feature mask $\bar{\bm{z}}$ is obtained by averaging $Z$ over this minibatch. Finally, a mask vector $\bm{m}$ is calculated by applying feature mask normalization.}
	\label{fig:fm-layer}
\end{figure}

The feature mask module (FM-module), denoted as $f_\text{FM}(\cdot)$, aims to generate \emph{one} fixed feature mask $\bm{m}\in\mathbb{R}_{\ge 0}^D$ from the entire dataset as: $\bm{m} = f_\text{FM}(X_\text{all})$. 
In general, as shown in Fig.~\ref{fig:fm-layer}, the FM-module consists of three submodules: \emph{i}) non-linear transformation (\emph{blue}); \emph{ii}) batch-wise attenuation (\emph{green}); \emph{iii}) feature mask normalization (\emph{pink}). 

\paragraph{Non-linear Transformation}Let $X\in \mathbb{R}^{B\times D}$ denote a minibatch of size $B$. The minibatch $X$ is mapped to $Z=[\bm{z}_1, \bm{z}_2, \cdots, \bm{z}_B]^\top\in \mathbb{R}^{B\times D}$ under a non-linear transformation. By using this module, the complex (non-linear) relations between different input features can be captured during the training procedure; i.e., the corresponding feature mask is learned by taking the relation of all input features into consideration and a reliable feature selection can be expected. This idea is similar to~\cite{gui2019afs}, in which the input samples are compressed into lower-dimensional representations and mapped back to a matrix having the same size as the input feature matrix. In this paper, as an example, we stack two fully connected layers to realize this non-linear transformation:
\begin{equation}
\label{eq:non-linear-transformation}
\bm{z_i} = W_2\cdot\phi(W_1\cdot \bm{x}_i + \bm{b}_1) + \bm{b}_2\ ,
\end{equation}
where $W_1\in\mathbb{R}^{E\times D}$, $W_2\in\mathbb{R}^{D\times E}$, $\bm{b}_1\in\mathbb{R}^{E}$ and $\bm{b}_2\in\mathbb{R}^{D}$ with $E<D$. $\phi(\cdot)$ is a nonlinear function such as $\tanh(\cdot)$. It should be noted that the non-linear transformation module is generic and thus not restricted to the example mentioned above. Stacking more layers or using other layer types (e.g., convolutional layers and variational autoencoding layers) is also permitted within this non-linear transformation module.

\paragraph{Batch-Wise Attenuation}Each row vector of the resulting $Z$ obtained from the non-linear transformation module depends on the corresponding input sample. This means that each sample $\bm{x}_i$ in the original minibatch has its own mapping $\bm{z}_i$ in $Z$. However, this is not desirable for feature selection since it is generally required that all samples in the given data should have the same significant features. Therefore, by explicitly averaging $Z$ over the minibatch, a fixed vector for all samples within a minibatch can be obtained at each training iteration. Specifically, for each minibatch, we calculate
\begin{equation}
\label{eq:average}
\bar{\bm{z}} = \frac{1}{B}\sum_{i=1}^{B}\bm{z}_i\ .
\end{equation}

\paragraph{Feature Mask Normalization}This submodule is proposed to normalize the resulting $\bar{\bm{z}} = [\bar{z}_1, \bar{z}_2, \dots, \bar{z}_D]^\top$. That is to say, the relative importance of different input features are considered during training. Consequently, the resulting feature importance scores (the entries of the final feature mask vector $\bm{m}$) are more reliable. In this work, inspired by~\cite{vaswani2017attention}, we use the $\mathit{softmax}(\cdot)$ function to realize this module:
\begin{equation}
\label{eq:normalization}
\bm{m} = \mathit{softmax}(\bar{\bm{z}})\text{, with } m_i = \frac{e^{\bar{z}_i}}{\sum_{j=1}^{D}e^{\bar{z}_j}}.
\end{equation}

The three functions defined in~\eqref{eq:non-linear-transformation},~\eqref{eq:average} and~\eqref{eq:normalization} are differentiable, so the FM-module can be integrated with an arbitrary neural network targeting the downstream learning tasks by multiplying $\bm{m}$ with each input sample element-wisely. Accordingly, the neural network for the downstream task is trained on $\bm{m}\odot \bm{x}_i$ rather than the original $\bm{x}_i$ with $i \in\{1, 2, \dots, B\}$ during each iteration. Thereby, the FM-module is jointly trained with the learning network $g(\cdot)$. Once the FM-module is trained, the final feature mask $\bm{m}$ can be separately calculated over the entire training dataset, i.e., by applying the trained FM-module to the whole dataset $X_\text{all}$ according to~\eqref{eq:non-linear-transformation},~\eqref{eq:average} and~\eqref{eq:normalization}. As a result, each entry $m_i$ of $\bm{m}$ indicates the importance of the corresponding $i$-th feature, i.e., the $i$-th column of the dataset matrix $X_\text{all}$. Finally, we can select the top-$K$ features based on the resulting feature importance scores $\bm{m}$. Note that in contrast to conventional attention mechanism, the calculated feature mask $\bm{m}$ remains fixed while solving the learning task with the selected input features.

%% file: exp.tex
\section{Experiments}
\label{sec:exp}
To evaluate the proposed FM-module, we carried out intensive experiments on five datasets from different domains, including images, texts and speech signals. Additionally, several other state-of-the-art deep-learning-based feature selection approaches were used as the reference methods for comparison.   

\subsection{Setup}
The proposed FM-module and other six reference methods were respectively trained on five datasets in both supervised and unsupervised manners. A learning network $g(\cdot)$ with the same architecture was used for all methods for a fair comparison, since all considered methods were embedded feature selection approaches and did not have special requirements on the architecture of the learning network. In particular, in all conducted experiments, $g(\cdot)$ had two dense layers, with 128 and 64 neurons, respectively. After each dense layer, LeakyReLU~\cite{maas2013rectifier} with a rate of 0.2 was used as the activation layer. Before the output layer, a Dropout layer~\cite{srivastava2014dropout} with a dropout rate of 0.3 was used to reduce overfitting. 
It should be noted that each individual experiment was repeated five times with different random seeds to eliminate the occasionality of the training procedure in neural networks. In the following, we only report the averaged results together with their standard deviations. 
Furthermore, to be consistent with the previous studies, the accuracy on the test set was used to measure the performance of a feature selection method in all experiments.

\subsubsection{Reference Methods}
In this work, we used the following six state-of-the-art approaches from literature as reference methods for comparison: \emph{i}) Attention-based Feature Selection (AFS)~\cite{gui2019afs}; \emph{ii}) Concrete Autoencoder (ConAE)~\cite{abid2019concrete}; \emph{iii}) CancelOut~\cite{borisov2019cancelout}; \emph{iv}) Binary Stochastic Filtering (BSF)~\cite{trelin2020binary}; \emph{v}) Deep Feature Selection (DFS)~\cite{li2016deep}; \emph{vi}) Feature Selection using Deep Neural Networks (FSDNN)~\cite{roy2015feature}. The key hyperparameters of all reference methods were initially set based on the original literature and individually optimized by us for the considered datasets by grid search on a hold-out validation subset (10\% of the training samples). Furthermore, we used all raw features (RawF) as a performance upper bound, and randomly selected features (RSF) as a lower bound. It should be noted that the usage of raw features indicates that we did not do any additional feature extraction on a given dataset, although the dataset itself may consist of data based on extracted features.

\begin{figure*}[!ht]
	\centering
	\includegraphics[width=\linewidth]{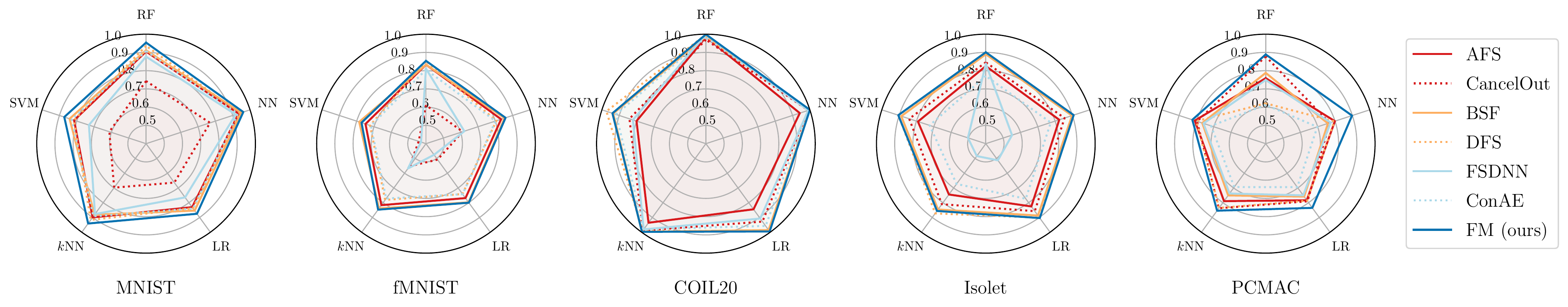}
	\caption{Downstream classification accuracy on five datasets using different classifiers. A larger area indicates better performance over one dataset in regard to different classifiers. Generally speaking, the FM-module (\emph{blue frame}) covers the largest area for all datasets.}
	\label{fig:all_res}
\end{figure*}

\subsubsection{Datasets}
Five well-known datasets, which frequently benchmark feature selection algorithms, were used in the conducted experiments, including three image datasets, MNIST~\cite{lecun1998gradient}, Fashion-MNIST (fMNIST)~\cite{xiao2017fashion} and COIL20~\cite{nene1996columbia}, one speech dataset Isolet~\cite{fanty1991spoken} and one text dataset PCMAC~\cite{li2017feature}. It should be noted that the latter three datasets consist of only limited number of samples. This is challenging for deep-learning-based approaches that typically require a large amount of data. Moreover, the Isolet and PCMAC datasets consist of extracted features. Detailed information of the datasets is listed in Table~\ref{tab:dataset}.
\begin{table}[ht]
	\caption{Overview of the datasets.}
	\label{tab:dataset}
	\renewcommand{\arraystretch}{1.25}
	\centering
		\begin{tabular}{cccccc}
			\hline
			\bfseries & \bfseries \# Features & \bfseries \# Train & \bfseries \# Test & \bfseries \# Classes & \bfseries Type \\
			\hline
			MNIST & 784 & 60000 & 10000 & 10 & Image\\
			fMNIST & 784 & 60000 & 10000 & 10 & Image\\
			COIL20 & 1024 & 1152 & 288 & 20 & Image\\
			Isolet & 617 & 1248 & 312 & 26 & Speech \\
			PCMAC  & 3289 & 1555 & 388 & 2 & Text \\
			\hline
	\end{tabular}
\end{table}

\subsubsection{Classifiers}
In order to verify whether the selected features are robust to different classifiers for the downstream learning tasks, the features selected by our method and the reference approaches were evaluated on different classifiers. Therefore, five well-known classifiers were trained on the selected features: \emph{i}) Random Forest (RF); \emph{ii}) Linear Support Vector Machine (SVM); \emph{iii}) $k$-Nearest Neighbor ($k$NN); \emph{iv}) Logistic Regression (LR); \emph{v}) Neural Network (NN). The five classifiers above were implemented using scikit-learn~\cite{scikit-learn} and TensorFlow~\cite{abadi2016tensorflow}. We followed the default or commonly used settings for all five classifiers in all experiments for a fair comparison.

\subsection{Basic Experiments}
The basic experiments aim to show whether the proposed FM-module can generally outperform other state-of-the-art feature selection methods. Fig.~\ref{fig:all_res} illustrates the performance over the five datasets with respect to the five different downstream classifiers, trained on the top-50 features selected by different methods. As can be seen, the FM-module outperforms other reference methods in 22 out of 25 cases, empirically showing that our method can select the most informative and representative features with robust performance on different classifiers. Due to the paper length limitation, we use the results on the Random Forest as an example (Table~\ref{tab:basic_exp}) for the rest part of this paper. In total, our method shows notably better performance than most methods. In particular, although DFS achieved moderately similar performance as our method in the conducted experiments, the FM-module does not possess that many hyperparameters as DFS: DFS uses $\ell_1$- and $\ell_2$-regularization in both its feature selection part and learning network. Consequently, four critical hyperparameters for the regularization terms have to be carefully fine-tuned in DFS. In contrast, our method does not introduce new loss terms and is thus easier to use. 

\begin{table}[ht]
	\renewcommand{\arraystretch}{2.2}
	\centering
	\caption{Downstream classification accuracy based on top-50 selected features in a supervised learning setup (random forest as the classifier).}
	\label{tab:basic_exp}
	\begin{tabular}{cccccc}
		\hline
		\bfseries & \bfseries MNIST & \bfseries fMNIST & \bfseries COIL20 & \bfseries Isolet & \bfseries PCMAC \\
		\hline
		RSF & \makecell[c]{0.842 \\ {\scriptsize($\pm$ 0.038)}}& \makecell[c]{0.828 \\ {\scriptsize($\pm$ 0.008)}}& \makecell[c]{0.991 \\ {\scriptsize($\pm$ 0.007)}}& \makecell[c]{0.835 \\ {\scriptsize($\pm$ 0.027)}}& \makecell[c]{0.584 \\ {\scriptsize($\pm$ 0.038)}}\\ 
		RawF & \makecell[c]{0.969 \\ {\scriptsize($\pm$ 0.000)}}& \makecell[c]{0.876 \\ {\scriptsize($\pm$ 0.000)}}& \makecell[c]{1.000 \\ {\scriptsize($\pm$ 0.000)}}& \makecell[c]{0.931 \\ {\scriptsize($\pm$ 0.000)}}& \makecell[c]{0.930 \\ {\scriptsize($\pm$ 0.000)}}\\ 
		\hline
		AFS & \makecell[c]{0.905 \\ {\scriptsize($\pm$ 0.022)}}& \makecell[c]{0.833 \\ {\scriptsize($\pm$ 0.001)}}& \makecell[c]{0.984 \\ {\scriptsize($\pm$ 0.009)}}& \makecell[c]{0.825 \\ {\scriptsize($\pm$ 0.002)}}& \makecell[c]{0.759 \\ {\scriptsize($\pm$ 0.061)}}\\ 
		CancelOut& \makecell[c]{0.743 \\ {\scriptsize($\pm$ 0.033)}}& \makecell[c]{0.611 \\ {\scriptsize($\pm$ 0.024)}}& \makecell[c]{0.970 \\ {\scriptsize($\pm$ 0.005)}}& \makecell[c]{0.846 \\ {\scriptsize($\pm$ 0.013)}}& \makecell[c]{0.878 \\ {\scriptsize($\pm$ 0.001)}}\\ 
		BSF & \makecell[c]{0.912 \\ {\scriptsize($\pm$ 0.018)}}& \makecell[c]{0.833 \\ {\scriptsize($\pm$ 0.012)}}& \makecell[c]{0.988 \\ {\scriptsize($\pm$ 0.010)}}& \makecell[c]{0.890 \\ {\scriptsize($\pm$ 0.019)}}& \makecell[c]{0.788 \\ {\scriptsize($\pm$ 0.025)}}\\ 
		DFS & \makecell[c]{0.936 \\ {\scriptsize($\pm$ 0.005)}}& \makecell[c]{0.852 \\ {\scriptsize($\pm$ 0.001)}}& \makecell[c]{0.998 \\ {\scriptsize($\pm$ 0.002)}}& \makecell[c]{0.900 \\ {\scriptsize($\pm$ 0.006)}}& \makecell[c]{0.621 \\ {\scriptsize($\pm$ 0.036)}}\\ 
		FSDNN & \makecell[c]{0.877 \\ {\scriptsize($\pm$ 0.011)}}& \makecell[c]{0.811 \\ {\scriptsize($\pm$ 0.011)}}& \makecell[c]{0.998 \\ {\scriptsize($\pm$ 0.002)}}& \makecell[c]{0.837 \\ {\scriptsize($\pm$ 0.054)}}& \makecell[c]{0.749 \\ {\scriptsize($\pm$ 0.014)}}\\ 
		ConAE & \makecell[c]{0.908 \\ {\scriptsize($\pm$ 0.022)}}& \makecell[c]{0.806 \\ {\scriptsize($\pm$ 0.017)}}& \makecell[c]{0.983 \\ {\scriptsize($\pm$ 0.007)}}& \makecell[c]{0.771 \\ {\scriptsize($\pm$ 0.011)}}& \makecell[c]{0.613 \\ {\scriptsize($\pm$ 0.000)}}\\ 
		FM (ours) & \makecell[c]{\textbf{0.954} \\ {\scriptsize($\pm$ 0.003)}}& \makecell[c]{\textbf{0.854} \\ {\scriptsize($\pm$ 0.003)}}& \makecell[c]{\textbf{0.999} \\ {\scriptsize($\pm$ 0.002)}}& \makecell[c]{\textbf{0.901} \\ {\scriptsize($\pm$ 0.013)}}& \makecell[c]{\textbf{0.889} \\ {\scriptsize($\pm$ 0.010)}}\\ 
		
		\hline
	\end{tabular}
\end{table}

\begin{figure}[!ht]
	\centering
	\includegraphics[width=1\linewidth]{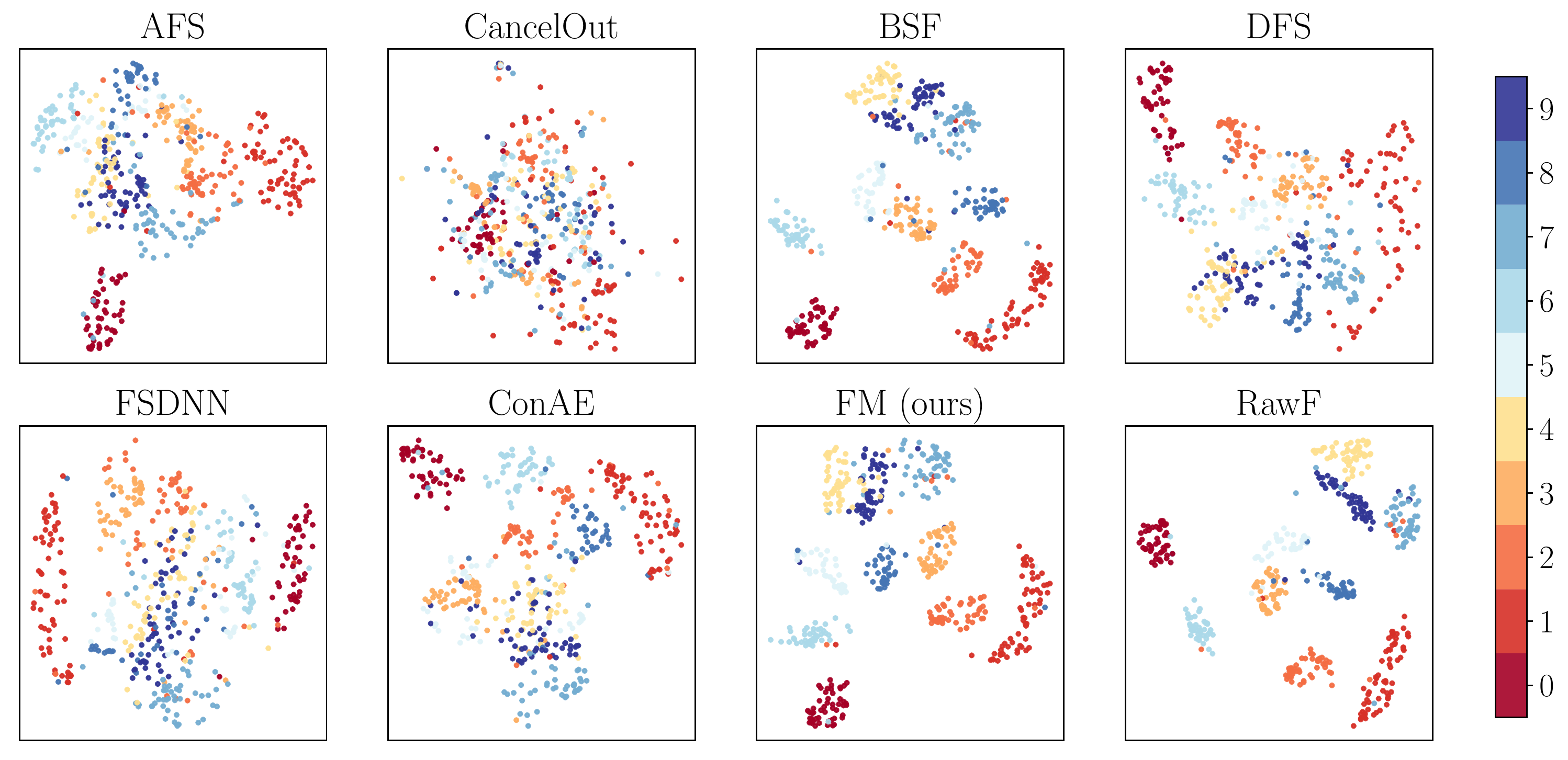}
	\caption{2D visualization of the MNIST dataset using the top-50 features (pixels). The selected features by the proposed FM-module well retain the data structure and perform similarly to the raw features.}
	\label{fig:2d_mnist_supervised}
\end{figure}

As mentioned before, in our experiment, RawF and RSF were used as the upper and lower bounds for evaluating feature selection methods, respectively. We argue that a carefully designed algorithm should have better performance than randomly selected features (RSF). Although previous studies often neglected the experiments with RSF, we surprisingly found that RSF led to better results than some carefully designed methods in some cases. For example, the method CancelOut has poorer performance than RSF on the three image datasets. Moreover, FSDNN had moderately poorer performance than RSF on the fMNIST datasets. In contrast, the FM-module significantly outperforms the RFS on all considered datasets. On the other hand, the FM-module and other reference methods have only small standard deviation with respect to different network initialization seeds. This is feasible and shows that all methods have consistent performance regarding different initialization. 

To have a more intuitive understanding, as shown in Fig.~\ref{fig:2d_mnist_supervised}, we used UMAP~\cite{mcinnes2018umap} to visualize the MNIST dataset based on the selected features. Apparently, the FM-module can select the most informative and representative features because the samples from the same class are well clustered even using the top-50 features only. Similar visualization results can be observed for BSF which performed well on the MNIST dataset. In contrast, the selected features by CancelOut led to a poor visualization result; i.e., all samples are in mixture without any clusters, which also matches the classification accuracy in Table~\ref{tab:basic_exp}.

\subsection{Performance over Different Numbers of Features}
\begin{figure}[ht]
	\centering
	\includegraphics[width=1\linewidth]{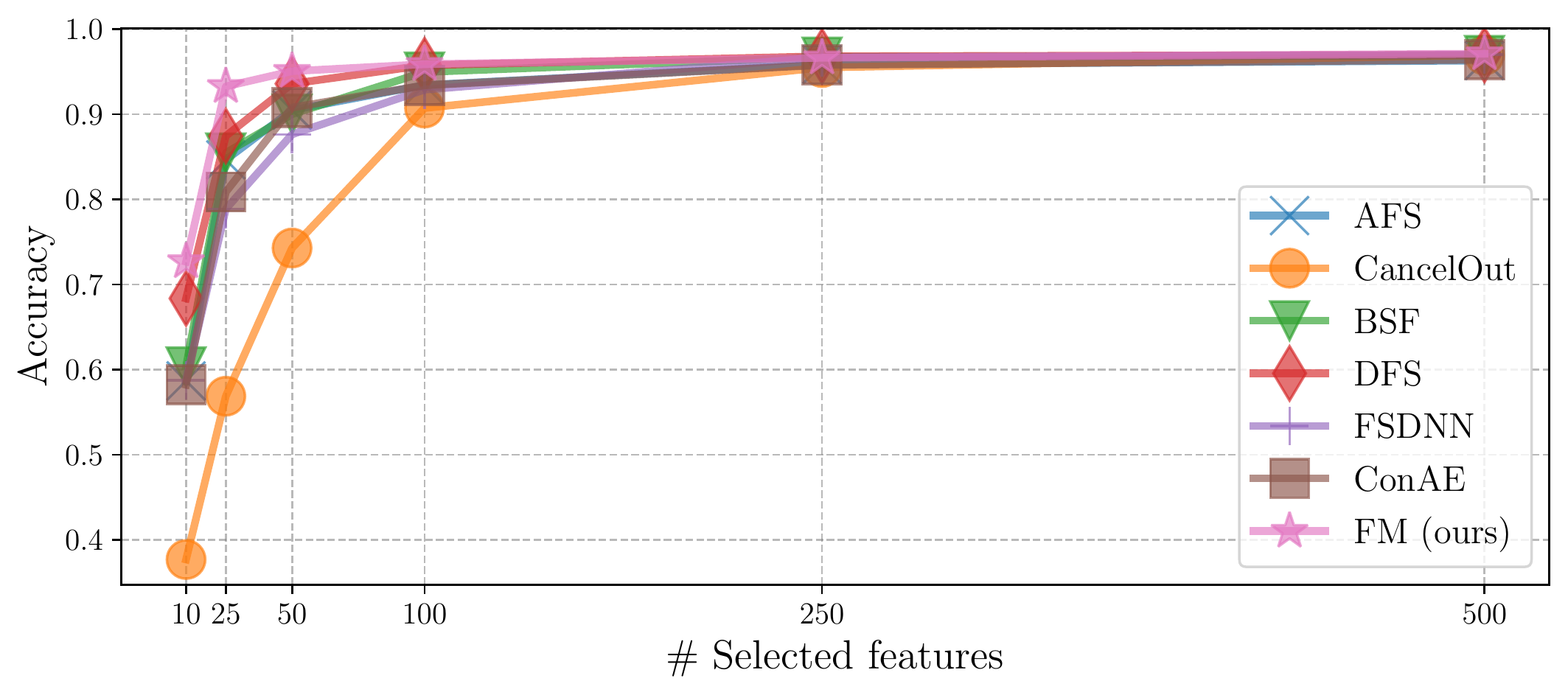}
	\caption{Classification accuracy obtained with different numbers of selected features on the MNIST dataset (random forest as the classifier).}
	\label{fig:accu_vs_num_fea}
\end{figure}

In practice, based on different requirements, it can be desirable to evaluate multiple feature subsets of different sizes. Thereby, different numbers of features were selected and compared in this subsection. Fig.~\ref{fig:accu_vs_num_fea} shows an exemplary experimental result on the MNIST dataset using the random forest as the classifier. Six different numbers of features were investigated: 10, 25, 50, 100, 250 and 500. They respectively correspond to 1.3\%, 3.2\%, 6.4\%, 12.8\%, 31.9\% and 63.8\% of all raw features (pixels) of the original dataset.

The overall results show that the proposed FM-module can select more informative features in comparison with the other six approaches with respect to different feature subset sizes. We can notice that, on MNIST, the features selected by the FM-module results in an accuracy above 90\% when using only about 3.2\% of all pixels of the original image data. Furthermore, by using FM-module, 100 selected features (only about 12.8\% of all features) can already lead to similar performance with that of all features. This is valuable for studying the salient features of a given dataset and reducing the data dimension more effectively and efficiently.

It should be also mentioned that ConAE has to be individually trained for each required feature subset size, because ConAE requires a predefined number of features as one of the training hyperparameters. In contrast, other methods, including our FM-module, do not require the number of target features as a parameter for training. In other words, all methods except for ConAE output a feature importance vector. Different numbers of features can thus be selected by choosing the features corresponding to the highest importance scores. This is more practical and efficient since training of neural networks is typically time consuming.

\subsection{Unsupervised Settings}
The proposed FM-module is capable of being jointly trained with different networks. Therefore, our method can be applied to unsupervised feature selection tasks as well. In this case, a typical learning objective is to minimize the reconstruction loss between the input and the reconstructed data. Thereby, the mean squared error (MSE) loss can be a canonical choice in that context. To build up the unsupervised learning environment, $g(\cdot)$ in Fig.~\ref{fig:network} becomes an autoencoder with three hidden dense layers using LeakyReLU as activation functions, having 128, 64 and 128 neurons, respectively. The hyperparameters of all methods were kept identical to those in the previous subsections.

\begin{table}[ht]
	\renewcommand{\arraystretch}{2.2}
	\centering
	\caption{Downstream classification accuracy based on the top-50 selected features in an unsupervised learning setup (Random forest as the classifier).}
	\label{tab:unsupervised}
	\begin{tabular}{cccccc}
		\hline
		\bfseries & \bfseries MNIST & \bfseries fMNIST & \bfseries COIL20 & \bfseries Isolet & \bfseries PCMAC \\
		\hline
		RSF & \makecell[c]{0.842 \\ {\scriptsize($\pm$ 0.038)}}& \makecell[c]{0.828 \\ {\scriptsize($\pm$ 0.008)}}& \makecell[c]{0.991 \\ {\scriptsize($\pm$ 0.007)}}& \makecell[c]{0.835 \\ {\scriptsize($\pm$ 0.027)}}& \makecell[c]{0.584 \\ {\scriptsize($\pm$ 0.038)}}\\ 
		RawF & \makecell[c]{0.969 \\ {\scriptsize($\pm$ 0.000)}}& \makecell[c]{0.876 \\ {\scriptsize($\pm$ 0.000)}}& \makecell[c]{1.000 \\ {\scriptsize($\pm$ 0.000)}}& \makecell[c]{0.931 \\ {\scriptsize($\pm$ 0.000)}}& \makecell[c]{0.930 \\ {\scriptsize($\pm$ 0.000)}}\\ 
		\hline
		AFS & \makecell[c]{0.883 \\ {\scriptsize($\pm$ 0.010)}}& \makecell[c]{0.808 \\ {\scriptsize($\pm$ 0.023)}}& \makecell[c]{0.970 \\ {\scriptsize($\pm$ 0.029)}}& \makecell[c]{0.759 \\ {\scriptsize($\pm$ 0.025)}}& \makecell[c]{0.548 \\ {\scriptsize($\pm$ 0.003)}}\\ 
		CancelOut& \makecell[c]{0.875 \\ {\scriptsize($\pm$ 0.018)}}& \makecell[c]{0.805 \\ {\scriptsize($\pm$ 0.020)}}& \makecell[c]{0.977 \\ {\scriptsize($\pm$ 0.019)}}& \makecell[c]{0.782 \\ {\scriptsize($\pm$ 0.037)}}& \makecell[c]{0.583 \\ {\scriptsize($\pm$ 0.021)}}\\ 
		BSF & \makecell[c]{0.869 \\ {\scriptsize($\pm$ 0.022)}}& \makecell[c]{0.803 \\ {\scriptsize($\pm$ 0.007)}}& \makecell[c]{0.950 \\ {\scriptsize($\pm$ 0.020)}}& \makecell[c]{0.833 \\ {\scriptsize($\pm$ 0.006)}}& \makecell[c]{0.556 \\ {\scriptsize($\pm$ 0.023)}}\\ 
		DFS & \makecell[c]{0.817 \\ {\scriptsize($\pm$ 0.123)}}& \makecell[c]{0.815 \\ {\scriptsize($\pm$ 0.019)}}& \makecell[c]{0.918 \\ {\scriptsize($\pm$ 0.099)}}& \makecell[c]{0.818 \\ {\scriptsize($\pm$ 0.046)}}& \makecell[c]{0.643 \\ {\scriptsize($\pm$ 0.103)}}\\ 
		FSDNN & \makecell[c]{0.935 \\ {\scriptsize($\pm$ 0.002)}}& \makecell[c]{0.815 \\ {\scriptsize($\pm$ 0.002)}}& \makecell[c]{0.943 \\ {\scriptsize($\pm$ 0.046)}}& \makecell[c]{0.825 \\ {\scriptsize($\pm$ 0.023)}}& \makecell[c]{0.667 \\ {\scriptsize($\pm$ 0.038)}}\\ 
		ConAE & \makecell[c]{0.868 \\ {\scriptsize($\pm$ 0.027)}}& \makecell[c]{0.808 \\ {\scriptsize($\pm$ 0.024)}}& \makecell[c]{0.976 \\ {\scriptsize($\pm$ 0.010)}}& \makecell[c]{0.790 \\ {\scriptsize($\pm$ 0.030)}}& \makecell[c]{0.575 \\ {\scriptsize($\pm$ 0.011)}}\\ 
		FM (ours)& \makecell[c]{\textbf{0.936} \\ {\scriptsize($\pm$ 0.009)}}& \makecell[c]{\textbf{0.842} \\ {\scriptsize($\pm$ 0.003)}}& \makecell[c]{\textbf{0.998} \\ {\scriptsize($\pm$ 0.002)}}& \makecell[c]{\textbf{0.865} \\ {\scriptsize($\pm$ 0.005)}}& \makecell[c]{\textbf{0.692} \\ {\scriptsize($\pm$ 0.033)}}\\ 
		\hline
	\end{tabular}
\end{table}

In particular, all methods were trained in an unsupervised manner and the resulting selected features were evaluated by classifiers targeting classification tasks. Table~\ref{tab:unsupervised} shows the results on the random forest classifier trained on the selected features. The FM-module achieved the best performance on all five datasets against the reference methods. This shows that the proposed method is flexible and can be applied to different learning environments, i.e., in both supervised and unsupervised learning. Moreover, it can be noticed that the selected features did not show significant performance difference or deterioration on MNIST, fMNIST and COIL20 from the supervised setup to the unsupervised setup. One plausible reason might be that the MSE loss function can also guide the feature selection algorithms to select similar informative features as the categorical cross entropy loss on the image datasets. However, when it comes to the speech or text datasets (Isolet and PCMAC), where the input features are sometimes sparse (containing many zeros), MSE as the learning objective becomes difficult to minimize. Hence, the performance on selected features is poorer than that obtained in a supervised learning setup.

\begin{figure}[!ht]
	\centering
	\includegraphics[width=1\linewidth]{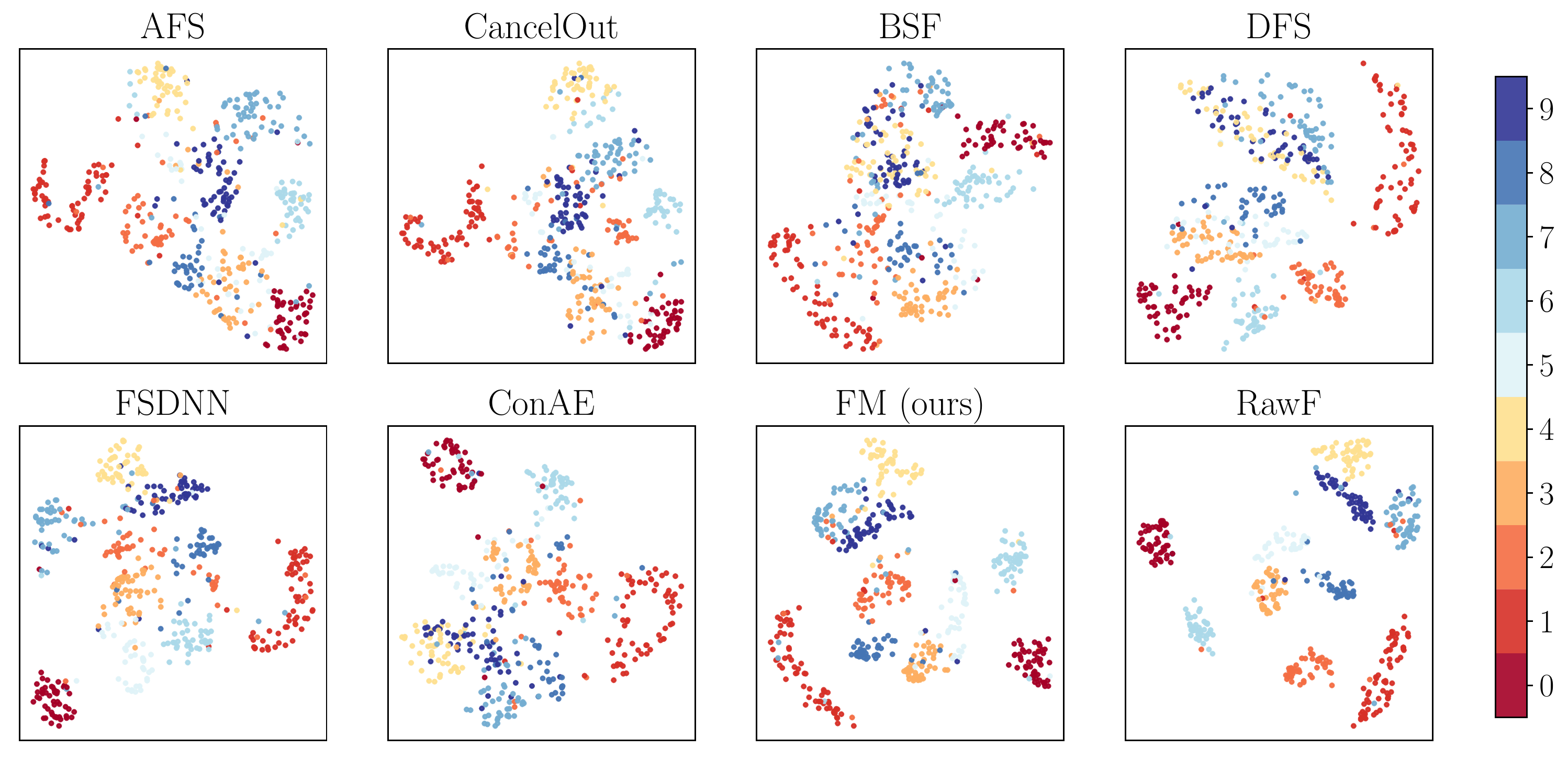}
	\caption{2D visualization of the MNIST dataset using the top-50 features (pixels) in an unsupervised setup.}
	\label{fig:2d_mnist_unsupervised}
\end{figure}

Moreover, Fig.~\ref{fig:2d_mnist_unsupervised} shows the 2D visualization of the MNIST dataset based on the top-50 selected features in an unsupervised setup. Notably, the FM-module successfully selected the representative features without relying on the label information and the samples can be well clustered as those with the raw features (RawF). On the contrary, many other reference methods failed to select the most informative features, leading to mixtures in the low-dimensional UMAP projection.

\subsection{Insight into the FM-module}
In comparison with other considered state-of-the-art approaches, the proposed FM-module enjoys a key property that it is free from special hyperparameters such as additional decays for $\ell_1$- or $\ell_2$-regularization. Nonetheless, this subsection investigates the advantages and limitations of the proposed FM-module to provide a deep insight into this method and its key components.

\subsubsection{A Visual Understanding of the FM-module}
\begin{figure}
	\centering
	\includegraphics[width=1\linewidth]{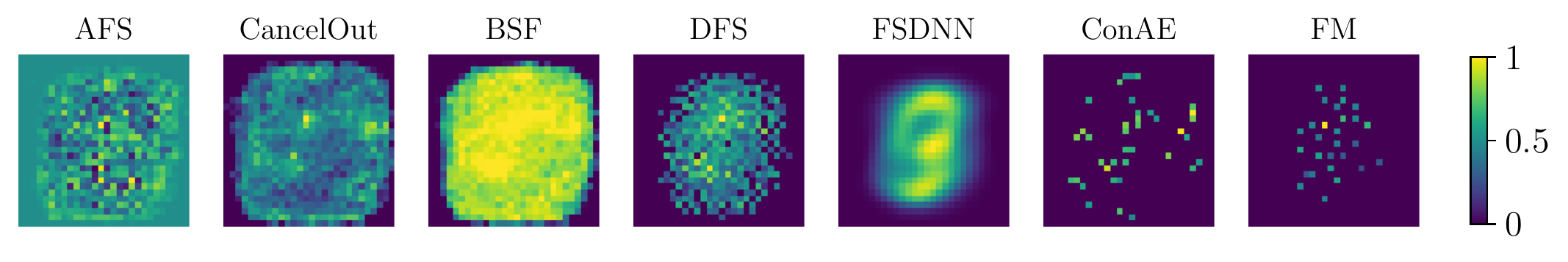}
	\caption{The visualization of the learned feature importance vectors for the reference methods and FM-module on the MNIST dataset.}
	\label{fig:fea_mask_mnist}
\end{figure}
\begin{figure}
	\centering
	\includegraphics[width=1\linewidth]{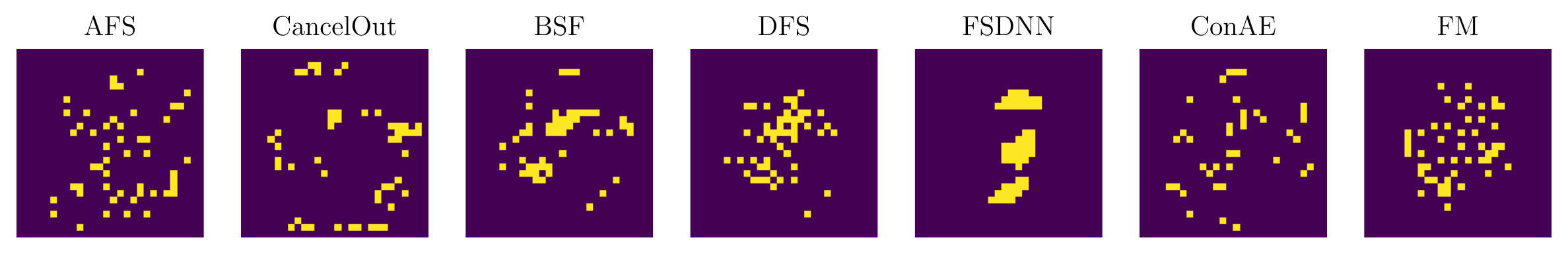}
	\caption{The selected top-50 features based on the learned feature importance vectors in a supervised setup on the MINIST dataset.}
	\label{fig:top50_fea_mnist}
\end{figure}
In order to gain a more intuitive understanding of the superiority of the proposed FM method, we visualize the learned feature importance vector of all reference methods as shown in Fig.~\ref{fig:fea_mask_mnist}. To make the comparison clearer, we re-scaled the resulting feature importance vectors into the range between 0 and 1, and reshaped them into $28\times 28$ for the MNIST dataset for better visualization. Apparently, the FM-module can select the most informative features, i.e., the pixels distributing in the center of images. Although DFS and FSDNN can also select the centered pixels, they often assign similar importance scores to neighboring pixels. In this case, redundant features (pixels) are selected and do not bring any additional information for the subsequent learning tasks. On the contrary, other methods, such as CancelOut and AFS, tend to assign high importance scores to the pixels at the edges of the image. However, the pixels around the edges have values of zero and do not provide any information for classification. Thereby, non-related features are mistakenly selected by these methods. In Fig.~\ref{fig:top50_fea_mnist}, we highlight the top-50 features based on the learned feature mask in Fig.~\ref{fig:fea_mask_mnist}. The visualization of the top-50 features confirms the observations made above. Interestingly, the number of unique features selected by ConAE is not equal to 50, because ConAE might repeatedly select the same feature during training.

\subsubsection{Ablation Study of the FM-module}
This subsection studies the necessity of the novel batch-wise attenuation (BA) and feature mask normalization (FMN) within the FM-module. Table~\ref{tab:ablation-fm} presents the overall results of the ablation study. Generally, both submodules contribute to the final performance (e.g., the classification accuracy on a random forest classifier). For example, the batch-wise attenuation has significant contribution when not using feature mask normalization (w/o FMN), while BA leads to limited performance improvement if using FMN. We argue that the utilization of FMN has two major advantages: One is the introduction of the non-negativity of the resulting feature mask due to the $\mathit{softmax}(\cdot)$ function; the other is the consideration of the relative importance among individual features due to the denominator of $\mathit{softmax}(\cdot)$. The latter property of FMN lacks of attention in previous studies despite its importance. On the other hand, the batch-wise attenuation forces that the feature mask vector to be same for all input samples, differentiating the proposed method from conventional attention mechanism. Using the same feature mask vector for all samples is more intuitive and reasonable for feature selection and shows better results in the conducted experiments.

\begin{table}[ht]
	\caption{Ablation study of the FM-module (exemplary classification accuracy on MNIST using a random forest classifier).}
	\label{tab:ablation-fm}
	\renewcommand{\arraystretch}{1.3}
	\centering
	\begin{tabular}{ccc}
		\hline
		\bfseries & \bfseries w/o FMN & \bfseries w/ FMN\\
		\hline
		\bfseries w/o BA & 0.720 \scriptsize($\pm$ 0.005) & 0.932 \scriptsize($\pm$ 0.005) \\ 
		\bfseries w/ BA & 0.898 \scriptsize($\pm$ 0.008) & 0.954 \scriptsize($\pm$ 0.003) \\ 
		\hline
	\end{tabular}
	
\end{table}

\subsubsection{Time Complexity}
This subsection compares the time complexity of reference methods and the FM-module. The batch size is denoted as $B$, the number of raw features is denoted as $D$, and the number of selected features is denoted as $K$. Furthermore, without loss of generality, the learning network $g(\cdot)$ is assumed to consist of fully connected layers, having $DN_1 + N_{others}$ parameters, where $N_1$ represents the number of neurons in the first hidden dense layer, and $N_{others}$ represents the number of all other parameters within this network. Moreover, the AFS and FM methods require the input data first to be transformed into a lower-dimensional representation with the dimension $E$. In the following, we omit the bias terms in all layers for simplicity.

\begin{figure}[ht]
	\centering
	\includegraphics[width=1\linewidth]{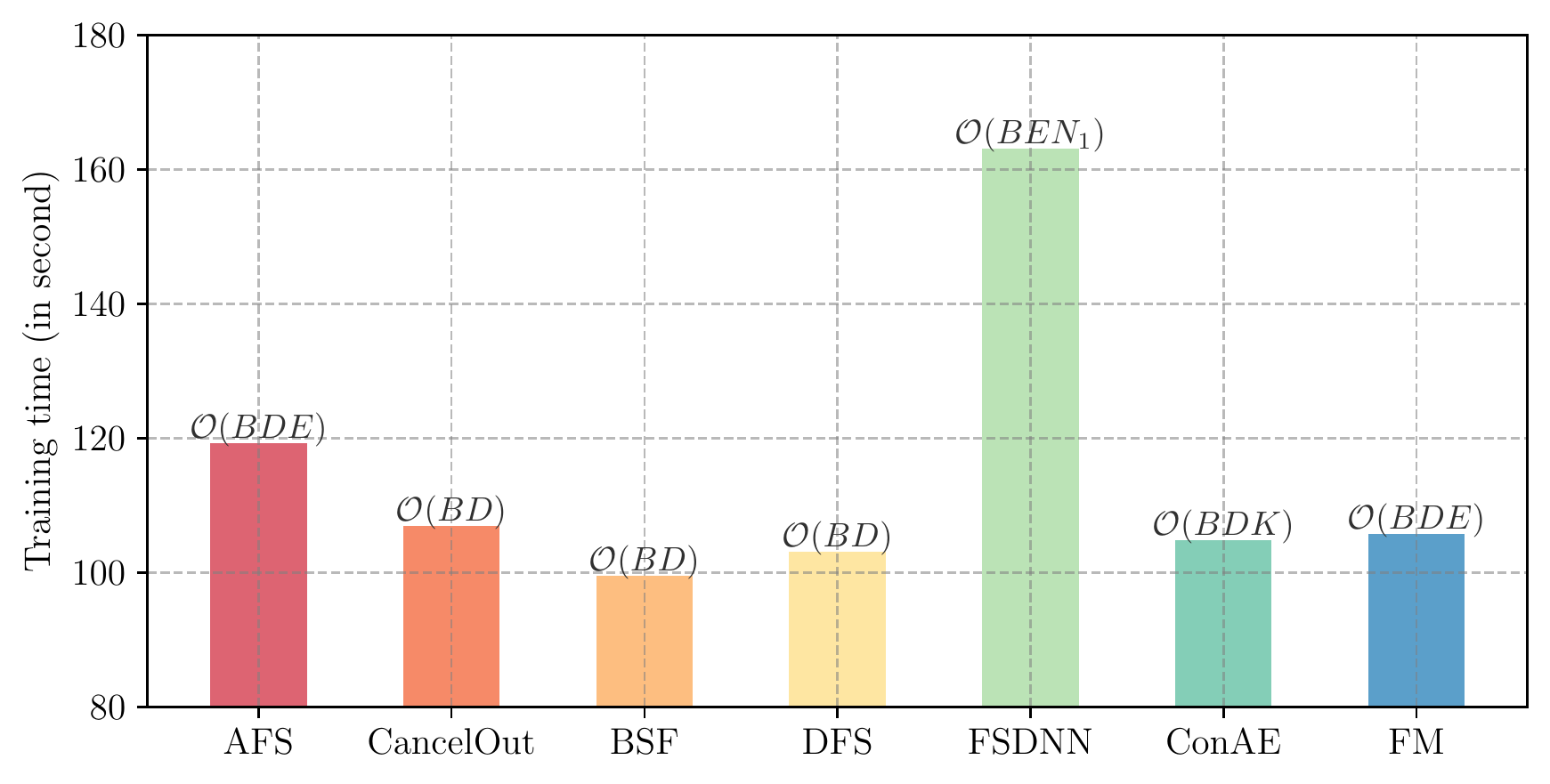}
	\caption{Time complexity and actual time consumption of the reference methods and FM-module.}
	\label{fig:complexity}
\end{figure}

Fig.~\ref{fig:complexity} shows the time complexity of all methods, where we omit the complexity of the learning network $g(\cdot)$. In other words, we only compared the time complexity of the special network or layer proposed to perform feature selection in each method. Generally, CancelOut, BSF and DFS have the lowest  time complexity $\mathcal{O}(BD)$ during training, because the initialized feature importance vector is directly element-wisely multiplied to the input data. On the contrary, AFS, ConAE and FM-module have a slightly higher time complexity due to the matrix multiplications. Furthermore, in AFS~\cite{gui2019afs}, the authors state that their method can be generalized to a version with deeper attention modules. In this case, the time complexity of AFS can be significantly larger than other methods, so we only report the minimal time complexity of AFS. 
Although the proposed method does not have the smallest time complexity, in practice, this gap can be omitted because the learning part $g(\cdot)$ typically has notably greater time complexity. Fig.~\ref{fig:complexity} shows the actual time consumption for the seven methods trained on MNIST with 100 epochs as an example\footnote{The implementation of FSDNN for generating feature importance vector was not based on GPU, so the actual time consumption was significantly larger than others.}. Apparently, only little time consumption difference can be observed among different methods. Furthermore, we argue that this time consumption difference can be even smaller when we turn to a deeper learning network $g(\cdot)$.

\subsubsection{Deeper FM-module}
As discussed before, stacking multiple layers can lead to a deeper version of the FM-module, i.e. $\bm{z}_i = f_1\circ f_2\circ \cdots \circ f_L(\bm{x}_i)$, where $f_i(\cdot)$ is one non-linear transformation layer, such as a dense layer with a $\tanh(\cdot)$ function. According to our experiment, this extension of the FM-module is not necessary. Specifically, we evaluated the performance of the FM-module with different numbers of hidden layers, ranging from one hidden layer (the proposed version) to six hidden layers. The resulting downstream classification accuracy on MNIST was between 0.950 to 0.959. This empirically shows that the current FM-module with one hidden layer (128 neurons) only; i.e., the structure defined in Subsection~\ref{subsec:fm-module} can guarantee a reliable performance and select the most important features. The reason might be that the current FM-module already has enough capacity to generate the feature mask from the data.

\subsubsection{Sensitivity to Initialization}
The initialization is an important issue in deep-learning-based feature selection approaches. For example, AFS and CancelOut require a careful initialization~\cite{gui2019afs,borisov2019cancelout}. To eliminate this concern, we justified our method on the following different initialization methods: \emph{i}) weights initialized from a uniform distribution; \emph{ii}) weights initialized from a normal distribution; \emph{iii}) weights initialized with constants (ones); \emph{iv}) weights initialized by using Xavier methods~\cite{glorot2010understanding}. The standard deviation of the final classification accuracy for different initialization methods was within 0.4\%. This empirically shows that our method is not sensitive to different initialization methods.

\subsubsection{Impact of Batch Size}
The proposed batch-wise attenuation calculates the average feature mask over a mini-batch during each training iteration. Hence, we evaluated the feature selection performance with respect to different batch sizes on MNIST: 4, 8, 16, 32, 64, 128, 256, 512 and 1024. Specifically, the model was trained for a fixed number of iterations with different batch sizes to obtain a fair comparison. Generally, the FM-module yields a good performance with different batch sizes in a wide range from 16 to 1024 with a classification accuracy ranging from 0.951 to 0.956. On the contrary, the notably small batch sizes (4 and 8) led to slightly worse performance of 0.930 and 0.947. Although the differences in performance are limited, we still observed a slight tendency that larger batch sizes could lead to better selection results. This matches our expectation because the batch-wise attenuation can generate more representative $\bar{\bm{z}}$ with larger batch sizes.

%% file: conclusion.tex
\section{Conclusion}
This paper proposes a novel feature mask module for feature selection in combination with neural networks leveraging the novel batch-wise attenuation and feature mask normalization. In comparison to existing deep-learning-based approaches, our method does not require additional loss terms in the overall learning objective, and the module can thus be easily trained with arbitrary networks in a joint fashion. Experiments show that the FM-module can be used for both supervised and unsupervised feature selection tasks. Additionally, the comprehensive experiments conducted on datasets from different domains demonstrate the effectiveness and superiority of our approach in comparison with other state-of-the-art methods. Further research might explore the possibility to leverage the FM-module to perform data selection and network reduction. Moreover, the authors plan to apply the FM-module to other scientific domains, such as gene analysis or semiconductor test data analysis.

%% file: IJCNN_main.bbl
\begin{thebibliography}{10}

\bibitem{li2017feature}
Jundong Li, Kewei Cheng, Suhang Wang, Fred Morstatter, Robert~P Trevino,
  Jiliang Tang, and Huan Liu.
\newblock Feature selection: A data perspective.
\newblock {\em ACM Computing Surveys (CSUR)}, 50(6):1--45, 2017.

\bibitem{li2016deep}
Yifeng Li, Chih-Yu Chen, and Wyeth~W Wasserman.
\newblock Deep feature selection: theory and application to identify enhancers
  and promoters.
\newblock {\em Journal of Computational Biology}, 23(5):322--336, 2016.

\bibitem{guyon2003introduction}
Isabelle Guyon and Andr{\'e} Elisseeff.
\newblock An introduction to variable and feature selection.
\newblock {\em Journal of machine learning research}, 3(Mar):1157--1182, 2003.

\bibitem{brown2012conditional}
Gavin Brown, Adam Pocock, Ming-Jie Zhao, and Mikel Luj{\'a}n.
\newblock Conditional likelihood maximisation: a unifying framework for
  information theoretic feature selection.
\newblock {\em The journal of machine learning research}, 13(1):27--66, 2012.

\bibitem{peng2005feature}
Hanchuan Peng, Fuhui Long, and Chris Ding.
\newblock Feature selection based on mutual information criteria of
  max-dependency, max-relevance, and min-redundancy.
\newblock {\em IEEE Transactions on pattern analysis and machine intelligence},
  27(8):1226--1238, 2005.

\bibitem{goodfellow2016deep}
Ian Goodfellow, Yoshua Bengio, and Aaron Courville.
\newblock {\em Deep learning}, volume~1.
\newblock 2016.

\bibitem{gui2019afs}
Ning Gui, Danni Ge, and Ziyin Hu.
\newblock Afs: An attention-based mechanism for supervised feature selection.
\newblock In {\em Proceedings of the AAAI Conference on Artificial
  Intelligence}, volume~33, pages 3705--3713, 2019.

\bibitem{wang2014attentional}
Qian Wang, Jiaxing Zhang, Sen Song, and Zheng Zhang.
\newblock Attentional neural network: Feature selection using cognitive
  feedback.
\newblock In {\em Advances in neural information processing systems}, pages
  2033--2041, 2014.

\bibitem{abid2019concrete}
Abubakar Abid, Muhammed~Fatih Balin, and James Zou.
\newblock Concrete autoencoders for differentiable feature selection and
  reconstruction.
\newblock In {\em Proceedings of the 36th International Conference on Machine
  Learning, PMLR}, 2019.

\bibitem{han2018autoencoder}
Kai Han, Yunhe Wang, Chao Zhang, Chao Li, and Chao Xu.
\newblock Autoencoder inspired unsupervised feature selection.
\newblock In {\em 2018 IEEE International Conference on Acoustics, Speech and
  Signal Processing (ICASSP)}, pages 2941--2945. IEEE, 2018.

\bibitem{borisov2019cancelout}
Vadim Borisov, Johannes Haug, and Gjergji Kasneci.
\newblock Cancelout: A layer for feature selection in deep neural networks.
\newblock In {\em International Conference on Artificial Neural Networks},
  pages 72--83. Springer, 2019.

\bibitem{trelin2020binary}
Andrii Trelin and Aleš Procházka.
\newblock Binary stochastic filtering: feature selection and beyond, 2020.

\bibitem{roy2015feature}
Debaditya Roy, K~Sri~Rama Murty, and C~Krishna Mohan.
\newblock Feature selection using deep neural networks.
\newblock In {\em 2015 International Joint Conference on Neural Networks
  (IJCNN)}, pages 1--6. IEEE, 2015.

\bibitem{zou2005regularization}
Hui Zou and Trevor Hastie.
\newblock Regularization and variable selection via the elastic net.
\newblock {\em Journal of the royal statistical society: series B (statistical
  methodology)}, 67(2):301--320, 2005.

\bibitem{vaswani2017attention}
Ashish Vaswani, Noam Shazeer, Niki Parmar, Jakob Uszkoreit, Llion Jones,
  Aidan~N Gomez, {\L}ukasz Kaiser, and Illia Polosukhin.
\newblock Attention is all you need.
\newblock In {\em Advances in neural information processing systems}, pages
  5998--6008, 2017.

\bibitem{maas2013rectifier}
Andrew~L Maas, Awni~Y Hannun, and Andrew~Y Ng.
\newblock Rectifier nonlinearities improve neural network acoustic models.
\newblock In {\em Proceedings of the 30th International Conference on Machine
  Learning (ICML)}, 2010.

\bibitem{srivastava2014dropout}
Nitish Srivastava, Geoffrey Hinton, Alex Krizhevsky, Ilya Sutskever, and Ruslan
  Salakhutdinov.
\newblock Dropout: a simple way to prevent neural networks from overfitting.
\newblock {\em The journal of machine learning research}, 15(1):1929--1958,
  2014.

\bibitem{lecun1998gradient}
Yann LeCun, L{\'e}on Bottou, Yoshua Bengio, and Patrick Haffner.
\newblock Gradient-based learning applied to document recognition.
\newblock {\em Proceedings of the IEEE}, 86(11):2278--2324, 1998.

\bibitem{xiao2017fashion}
Han Xiao, Kashif Rasul, and Roland Vollgraf.
\newblock Fashion-mnist: a novel image dataset for benchmarking machine
  learning algorithms.
\newblock {\em arXiv preprint arXiv:1708.07747}, 2017.

\bibitem{nene1996columbia}
Sameer~A Nene, Shree~K Nayar, Hiroshi Murase, et~al.
\newblock Columbia object image library (coil-100).
\newblock 1996.

\bibitem{fanty1991spoken}
Mark Fanty and Ronald Cole.
\newblock Spoken letter recognition.
\newblock In {\em Advances in Neural Information Processing Systems}, pages
  220--226, 1991.

\bibitem{scikit-learn}
F.~Pedregosa, G.~Varoquaux, A.~Gramfort, V.~Michel, B.~Thirion, O.~Grisel,
  M.~Blondel, P.~Prettenhofer, R.~Weiss, V.~Dubourg, J.~Vanderplas, A.~Passos,
  D.~Cournapeau, M.~Brucher, M.~Perrot, and E.~Duchesnay.
\newblock Scikit-learn: Machine learning in {P}ython.
\newblock {\em Journal of Machine Learning Research}, 12:2825--2830, 2011.

\bibitem{abadi2016tensorflow}
Mart{\'\i}n Abadi, Paul Barham, Jianmin Chen, Zhifeng Chen, Andy Davis, Jeffrey
  Dean, Matthieu Devin, Sanjay Ghemawat, Geoffrey Irving, Michael Isard, et~al.
\newblock Tensorflow: A system for large-scale machine learning.
\newblock In {\em 12th $\{$USENIX$\}$ symposium on operating systems design and
  implementation ($\{$OSDI$\}$ 16)}, pages 265--283, 2016.

\bibitem{mcinnes2018umap}
Leland McInnes, John Healy, and James Melville.
\newblock Umap: Uniform manifold approximation and projection for dimension
  reduction.
\newblock {\em arXiv preprint arXiv:1802.03426}, 2018.

\bibitem{glorot2010understanding}
Xavier Glorot and Yoshua Bengio.
\newblock Understanding the difficulty of training deep feedforward neural
  networks.
\newblock In {\em Proceedings of the thirteenth international conference on
  artificial intelligence and statistics}, pages 249--256, 2010.

\end{thebibliography}
